\documentclass[conference]{IEEEtran}
\IEEEoverridecommandlockouts
\UseRawInputEncoding
\usepackage{cite}
\usepackage{amsmath,amssymb,amsfonts}
\usepackage{algorithmic}
\usepackage{graphicx}
\usepackage{textcomp}
\usepackage{tabularx}
\usepackage{xcolor}
\usepackage{placeins}
\usepackage{array}
\usepackage{enumitem}
\usepackage{hyperref}
\usepackage{multirow}
\def\BibTeX{{\rm B\kern-.05em{\sc i\kern-.025em b}\kern-.08em
    T\kern-.1667em\lower.7ex\hbox{E}\kern-.125emX}}

\usepackage{geometry}
\begin{document}

\title{Exponentially Weighted Instance-Aware Repeat Factor Sampling for Long-Tailed Object Detection Model Training in Unmanned Aerial Vehicles Surveillance Scenarios
 \\
\thanks{Corresponding Author: Abhishek Kumar. Email: abhishek.k.kumar@jyu.fi}

}

\author{\IEEEauthorblockN{Taufiq Ahmed$^{\star}$, Abhishek Kumar$^{\dagger}$, Constantino {\'A}lvarez Casado$^{\star}$, Anlan Zhang$^{\mathsection}$, \\ Tuomo H{\"a}nninen$^{\star}$, Lauri Loven$^{\star}$, Miguel Bordallo  L{\'o}pez$^{\star}$, Sasu Tarkoma$^{\mathparagraph\star}$}
\IEEEauthorblockA{\textit{$^{\star}$University of Oulu, Oulu, Finland} \\
\textit{$^{\dagger}$University of Jyv\"{a}skyl\"{a}, Jyv\"{a}skyl\"{a}, Finland} \\
\textit{$^{\mathparagraph}$University of Helsinki, Helsinki, Finland} \\ 
\textit{$^{\mathsection}$University of Southern California, Los Angeles, United States}
}
}

\maketitle

\begin{abstract}
Object detection models often struggle with class imbalance, where rare categories appear significantly less frequently than common ones. Existing sampling-based rebalancing strategies, such as Repeat Factor Sampling (RFS) and Instance-Aware Repeat Factor Sampling (IRFS), mitigate this issue by adjusting sample frequencies based on image and instance counts. However, these methods are based on linear adjustments, which limit their effectiveness in long-tailed distributions. This work introduces Exponentially Weighted Instance-Aware Repeat Factor Sampling (E-IRFS), an extension of IRFS that applies exponential scaling to better differentiate between rare and frequent classes. E-IRFS adjusts sampling probabilities using an exponential function applied to the geometric mean of image and instance frequencies, ensuring a more adaptive rebalancing strategy. We evaluate E-IRFS on a dataset derived from the Fireman-UAV-RGBT Dataset and four additional public datasets, using YOLOv11 object detection models to identify fire, smoke, people and lakes in emergency scenarios. The results show that E-IRFS improves detection performance by 22\% over the baseline and outperforms RFS and IRFS, particularly for rare categories. The analysis also highlights that E-IRFS has a stronger effect on lightweight models with limited capacity, as these models rely more on data sampling strategies to address class imbalance. The findings demonstrate that E-IRFS improves rare object detection in resource-constrained environments, making it a suitable solution for real-time applications such as UAV-based emergency monitoring. The code is available at: \url{https://github.com/futurians/E-IRFS}.

\end{abstract}

\begin{IEEEkeywords}
Object detection, class imbalance, rebalancing strategies, instance-aware sampling, long-tailed distributions, UAV-based monitoring, emergency detection, YOLOv11
\end{IEEEkeywords}

%
%
\section{Introduction}

Effective object detection is essential in applications such as wildfire monitoring and emergency response \cite{Kularatne2024FireManPaper}. UAV-based monitoring extends the capability of fixed-camera networks by providing flexible aerial perspectives, enhancing situational awareness in dynamic and remote environments. In wildfire scenarios, timely detection of fire, smoke, and affected individuals is crucial for an effective response. However, deploying object detection models on UAVs introduces computational constraints, requiring lightweight models that can operate efficiently on resource-limited edge devices. This highlights the need for training strategies that enable compact models to achieve competitive detection performance.

\begin{figure}[t]
    \centering
    \includegraphics[width=\linewidth]{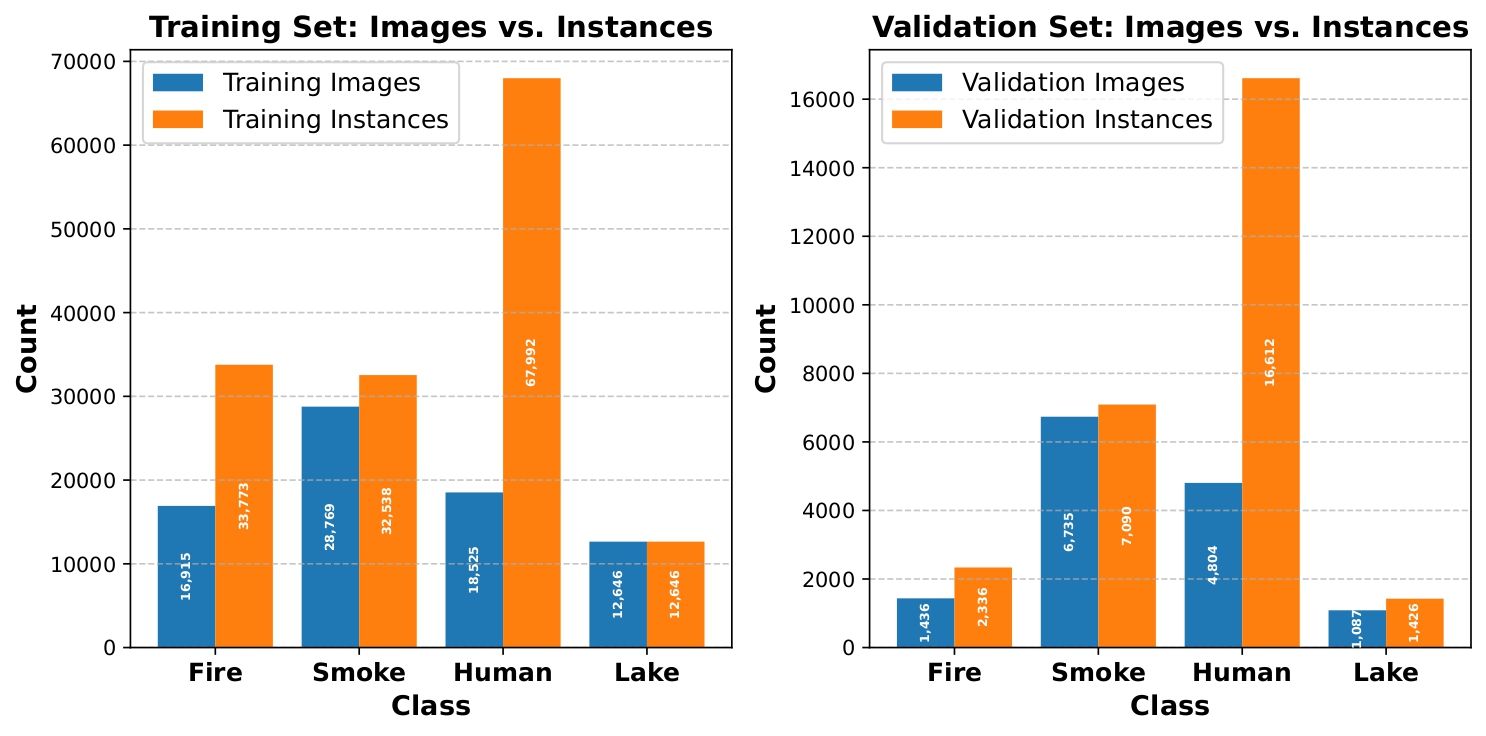}
    \vspace{-5mm}
    \caption{Class distribution of the training and validation sets in the custom benchmark dataset. Bars represent the number of images and instances per class, with vertically aligned counts. The imbalance highlights the dataset’s challenge for object detection models.}
    \label{fig:surveillance}
    \vspace{-3mm}
\end{figure}

Training object detection models for such applications is challenging due to the scarcity of representative datasets and the presence of class imbalance, where critical categories (e.g., fire, people in distress) appear far less frequently than common ones (e.g., buildings, vehicles) \cite{Elasal2018Frame,Farooq2023Synthetic}. This imbalance affects model learning, leading to biased predictions toward frequent classes and reduced accuracy for rare ones. Addressing this issue is essential in applications where the detection of rare objects is critical to decision making. Sampling-based methods, such as Repeat Factor Sampling (RFS) \cite{gupta2019lvis}, mitigate class imbalance by increasing the probability of selecting images that contain rare objects. Instance-Aware Repeat Factor Sampling (IRFS) improves upon RFS by incorporating both the image count and the instance count when determining the sampling weights \cite{yaman2023instance}.

In this context, this work introduces Exponentially Weighted Instance-Aware Repeat Factor Sampling (E-IRFS), a rebalancing strategy for improving rare-class detection in long-tailed object detection datasets. Unlike linear adjustments in existing methods, E-IRFS applies an exponential scaling mechanism to the geometric mean of image and instance frequencies, amplifying rare-class sampling while maintaining stable training. The evaluation is conducted on a custom dataset based on the Fireman-UAV-RGBT Dataset \cite{Kularatne2024FireManPaper,Kularatne2024firemandatabase} and four publicly available small datasets, covering fire, smoke, people, and lakes in emergency scenarios. Figure~\ref{fig:surveillance} illustrates the class distribution, emphasizing the disparity in image and instance counts. The method is compared against RFS, IRFS, and a baseline without rebalancing using YOLOv11 object detection models. The main contributions of this work are:

\begin{itemize} 
    \item The introduction of E-IRFS, an extension of IRFS that applies an exponential weighting mechanism to improve sampling-based rebalancing for rare classes. 
    \item A detailed evaluation of E-IRFS on an emergency response dataset, showing its effectiveness in improving detection accuracy for underrepresented categories. 
    \item An analysis of how E-IRFS benefits lightweight object detection models, where rebalancing strategies have a stronger effect due to the model’s limited capacity. 
\end{itemize}

These contributions refine class imbalance handling in object detection. E-IRFS enhances rare-class representation without excessive oversampling, benefiting lightweight models that struggle with imbalance. This makes it well-suited for UAV-based monitoring and edge applications. Table~\ref{tab:symbols} summarizes the symbols and notations used throughout the paper for consistency. \textcolor{black}{The code for E-IRFS and the implementation details of the custom benchmark dataset are available at: \url{https://github.com/futurians/E-IRFS}.}

\vspace{-2mm}
\begin{table}[ht!]

\centering
\caption{Summary of Symbols and Notations}
\vspace{-3mm}
\label{tab:symbols}
\setlength{\tabcolsep}{1.4em}
\def\arraystretch{1.1}
\begin{tabular}{|c|l|}
\hline
\textbf{Symbol} & \textbf{Definition} \\ 
\hline
\( c \) & Class index \\ 
\( i \) & Image index \\ 
\( f_c \) & Fraction of images containing class \( c \) \\ 
\( f_{i,c} \) & Fraction of images with at least one instance of \( c \) \\ 
\( f_{b,c} \) & Fraction of bounding boxes belonging to \( c \) \\ 
\( r_c \) & Repeat factor for class \( c \) \\ 
\( r_i \) & Repeat factor for image \( i \) \\ 
\( t \) & Threshold for oversampling activation \\ 
\( \alpha \) & Scaling parameter for E-IRFS \\ 
\( p_i \) & Probability of selecting image \( i \) for training \\ 
\hline
\end{tabular}
\end{table}

%
%
\section{Related Work}

Class imbalance in object detection datasets affects model performance by introducing biases toward frequent classes while reducing the ability to detect underrepresented objects. Many real-world datasets, particularly in autonomous systems, surveillance, and emergency response, follow long-tailed distributions where rare but critical categories appear infrequently \cite{Elasal2018Frame,Farooq2023Synthetic}. This imbalance is especially problematic in safety-critical applications, where missing rare objects can impact decision-making. Studies have shown that standard training pipelines often neglect underrepresented classes due to conventional loss formulations and the dominance of frequent classes in training data \cite{Gu2023systematic}. The issue is more pronounced in single-stage detectors, where class distribution affects anchor assignments and training dynamics, whereas two-stage detectors demonstrate greater robustness but still require adjustments to handle skewed datasets. Approaches to address class imbalance in object detection can be categorized into data-level methods, algorithm-level modifications, and hybrid techniques. Data-level methods adjust the dataset distribution by modifying how samples are selected or generated during training. Algorithm-level modifications introduce adjustments to the model architecture or loss function to enhance learning for underrepresented classes. Hybrid methods combine both strategies, integrating data augmentation with reweighting techniques to improve representation.

\subsection{Data Augmentation for Imbalance Mitigation}
Among data-level strategies, data augmentation is commonly used to artificially increase the number of rare-class samples. Techniques such as copy-paste augmentation \cite{ghiasi2021simple} insert rare objects into new images to enhance their representation. Similarly, Mosaic augmentation \cite{zhang2021mosaicos} merges multiple images into a single training sample to increase the variability of object occurrences. While data augmentation methods improve the diversity of training samples, they do not directly correct class distribution imbalances. Augmented samples remain subject to the same class frequencies as the original dataset, which means that rare classes may still be underrepresented during model training.

\subsection{Sampling-Based Methods for Rebalancing}

Sampling-based rebalancing adjusts the selection frequency of training samples without modifying the dataset itself. By increasing the presence of rare classes in training batches, these methods help mitigate class imbalance. Unlike loss reweighting, which alters optimization, they directly influence sample distribution. Early methods, such as class-balanced sampling, assigned equal probabilities to all classes but often led to overfitting due to excessive rare-class upsampling. More refined strategies constrain class occurrence to maintain stability. Among them, RFS \cite{gupta2019lvis} and IRFS \cite{yaman2023instance} have shown effectiveness. 


\subsubsection{Repeat Factor Sampling (RFS)} Repeat Factor Sampling (RFS) \cite{gupta2019lvis} was introduced to rebalance long-tailed object detection datasets by assigning higher sampling probabilities to images containing rare classes. The method defines a repeat factor based on the number of images in which a class appears, ensuring that classes with lower frequency are oversampled during training. For a given category \( c \), the category-level repeat factor is defined as:
\begin{equation}
    r_c = \max \left( 1, \sqrt{\frac{t}{f_c}} \right)
\end{equation}

For each image \( i \), the image-level repeat factor is computed as \( r_i = \max_{c \in i} r_c\), where \( c \in i \) represents the categories in image \( i \). Images are repeated during training based on \( r_i \), increasing the presence of rare-class instances. However, RFS only considers image-level frequency, ignoring instance counts, which can lead to inconsistencies when classes with similar image frequencies have different instance distributions.


\subsubsection{Instance-Aware Repeat Factor Sampling (IRFS)} To address the previous issue, IRFS method \cite{yaman2023instance} extends RFS by incorporating both image count and instance count into the sampling process. The method defines the repeat factor based on the geometric mean of these two values, allowing the sampling strategy to reflect both the presence and density of objects in the dataset.  The repeat factor for a given class \( c \) is defined as:
\begin{equation}
r_c = \max \left( 1, \sqrt{\frac{t}{\sqrt{f_{i,c} \cdot f_{b,c}}}} \right),
\end{equation}
determining how frequently images containing class \( c \) should be sampled during training, \( t \) is a predefined threshold that controls the degree of rebalancing by determining how aggressively rare classes should be oversampled. \( \sqrt{f_{i,c} \cdot f_{b,c}} \) is the geometric mean of the image frequency and instance frequency, ensuring a balanced measure of class presence and density.

%
%
\section{Proposed Methodology}

Existing sampling-based rebalancing techniques, such as RFS and IRFS, have improved rare-class representation in long-tailed object detection. However, these methods rely on linear weighting functions, which do not sufficiently amplify the presence of extremely rare classes. IRFS improves upon RFS by incorporating both image count and instance count, but its geometric mean formulation still applies only a limited adjustment for highly underrepresented categories. To address this limitation, we introduce Exponentially Weighted Instance-Aware Repeat Factor Sampling (E-IRFS), which applies an exponential weighting function to the geometric mean of image and instance frequencies. This adjustment ensures that rare classes receive a stronger sampling adjustment while preventing excessive overrepresentation of frequent classes. The exponential function enhances differentiation between rare and frequent categories, making the sampling process more adaptive to highly skewed distributions.

\subsection{Exponentially Weighted Instance-Aware Repeat Factor Sampling (E-IRFS)}
To further refine instance-aware sampling, we propose Exponentially Weighted Instance-Aware Repeat Factor Sampling (E-IRFS). This method amplifies the oversampling strength for rare classes by applying exponential scaling to the repeat factor. The new class-level repeat factor is formulated as:
\begin{equation}
    r_c = \exp \left( \alpha \cdot \sqrt{\frac{t}{\sqrt{f_{i,c} \cdot f_{b,c}}}} \right),
\end{equation}
where \( \alpha \) is a scaling parameter that controls the aggressiveness of sampling adjustments,  \( \exp(\cdot) \) ensures that small values of \( f_{i,c} \cdot f_{b,c} \) result in exponentially larger repeat factors, and the square root mean \( \sqrt{f_{i,c} \cdot f_{b,c}} \) balances both image and instance counts.

\subsection{Image-Level Sampling Using E-IRFS}
Once the category-level repeat factors \( r_c \) are computed, the image-level repeat factor \( r_i \) is given by \(r_i = \max_{c \in i} r_c\), where \( r_i \) determines how often an image is repeated during training. Higher \( r_i \) values indicate more frequent sampling of images containing rare objects.  The final sampling probability of selecting each image for training is given by equation \ref{eq:sampprob}, ensuring that rare classes receive higher sampling priority to counteract class imbalance:
\begin{equation}
    \label{eq:sampprob}
    p_i = \frac{r_i}{\sum_{j} r_j},
    \vspace{-2mm}
\end{equation}
where \( \sum_{j} r_j \) computes the total sum of all image-level repeat factors across the entire dataset and \( p_i \) represents the normalized probability of selecting image ii for training. This ensures that the sampling probabilities sum to 1, maintaining training stability while emphasizing rare categories effectively.

%
%
\subsection{Theoretical Analysis}
This section examines the properties of E-IRFS, including its scaling behavior, impact on class distribution, and effect on training convergence.


\subsubsection{Scaling Behavior and Growth Rate}
The effectiveness of sampling-based rebalancing depends on how it adjusts the sampling probability for rare classes. RFS and IRFS follow sublinear growth patterns, providing limited reinforcement for underrepresented categories. E-IRFS introduces an exponential scaling mechanism that increases the sampling probability more aggressively for rare classes while maintaining stability for frequent ones. The repeat factor \( r_c \) determines how strongly rare classes are oversampled. In RFS, it follows an inverse square root relationship with image frequency \( f_{i,c} \), while IRFS incorporates instance frequency \( f_{b,c} \) using a geometric mean:

\begin{enumerate}[label=\alph*)]
\vspace{2mm}
    \item RFS Growth Rate: \( r_c \propto f_{i,c}^{-1/2}  \)
    \vspace{3mm}
    \item IRFS Growth Rate: \( r_c \propto (f_{i,c} \cdot f_{b,c})^{-1/4}  \)
    \vspace{3mm}
    \item E-IRFS Growth Rate: \( r_c \propto \exp \left( \alpha \cdot (f_{i,c} \cdot f_{b,c})^{-1/4} \right)  \)
    \vspace{2mm}
\end{enumerate}

E-IRFS extends IRFS by applying an exponential function to the geometric mean, increasing the contrast between rare and frequent categories. Since the exponential function grows faster than polynomial functions, this approach increases model exposure to underrepresented categories while maintaining stable sampling for moderately frequent ones.


\subsubsection{Sampling Distribution and Training Balance}

E-IRFS amplifies the sampling probability of images containing rare objects, redistributing the probability mass to improve model exposure to underrepresented categories. The probability of selecting an image follows equation \ref{eq:sampprob}, where the exponential weighting enhances the inclusion of rare-class instances while preserving representation of frequent categories. The resulting class distribution in training is given by \( P_{\text{train}}(c) \propto r_c \cdot P_{\text{data}}(c) \), where \( P_{\text{data}}(c) \) is the original dataset distribution. For long-tailed datasets, \( P_{\text{data}}(c) \) follows a power-law distribution~\cite{clauset2009power}: \( P_{\text{data}}(c) \approx c_0 f_{i,c}^{-\gamma} \) (for some constant $c_0$ and exponent $\gamma$). For different sampling strategies, the training distribution is as follows:

\begin{enumerate}[label=\alph*)]
\vspace{1mm}
    \item RFS: \( P_{\text{train}}(c) \propto f_{i,c}^{-(\gamma + 1/2)}  \)
    \vspace{2mm}
    \item IRFS: \( P_{\text{train}}(c) \propto (f_{i,c} \cdot f_{b,c})^{-(\gamma + 1/4)}  \)
    \vspace{2mm}
    \item E-IRFS: \( P_{\text{train}}(c) \propto \exp \left( \alpha \cdot (f_{i,c} \cdot f_{b,c})^{-1/4} \right) f_{i,c}^{-\gamma}  \)
    \vspace{2mm}
\end{enumerate}

Since the exponential function in E-IRFS grows faster than polynomial adjustments, it significantly enhances the presence of rare classes in training. This ensures that underrepresented categories receive stronger sampling reinforcement while maintaining balance across all classes.

\subsubsection{Computational Complexity}
E-IRFS introduces an additional exponential computation and a geometric mean calculation per class. However, these operations remain in \( \mathcal{O}(N) \) per epoch, where \( N \) is the number of training images, similar to RFS and IRFS. Since the sampling process itself remains unchanged, the added computational overhead is minimal.

\subsubsection{Effect on Convergence and Sampling Stability}
In long-tailed object detection, rare classes contribute weaker gradient updates, making them harder to learn. E-IRFS mitigates this by increasing their representation, balancing gradient updates, and stabilizing training. This is especially beneficial for lightweight models with limited capacity to capture long-tailed distributions. The stability of this adjustment is supported by the mathematical properties of \( r_c \), with its first derivative given by:
\begin{equation} 
\frac{d}{d f_{i,c}} r_c = \alpha \cdot \exp \left( \alpha \cdot \sqrt{\frac{t}{\sqrt{f_{i,c} \cdot f_{b,c}}}} \right) \cdot \left( -\frac{t}{2\sqrt{f_{b,c} f_{i,c}^3}} \right) 
\end{equation}

Since the exponential function is always positive and the second term is negative,\( r_c \) is monotonically decreasing. This ensures that as the image frequency \( f_{i,c} \) decreases, the sampling rate increases, reinforcing rare-class representation in training. Furthermore, the second derivative is as follows:
\begin{equation} 
\frac{d^2}{d f_{i,c}^2} r_c = \alpha \cdot \exp \left( \alpha \cdot \sqrt{\frac{t}{\sqrt{f_{i,c} \cdot f_{b,c}}}} \right) \cdot \left( \frac{3t}{4\sqrt{f_{b,c} f_{i,c}^5}} \right) 
\end{equation}

Since this expression is always positive, \( r_c \) is convex. This ensures a smooth transition between frequent and rare categories, preventing abrupt sampling shifts and maintaining stable training dynamics while improving detection of underrepresented classes.

\subsection{Benchmark Database}

We created a custom benchmark for object detection by merging five different datasets (\cite{Kularatne2024FireManPaper}, \cite{forest-fire-and-smoke-rf4pd_dataset}, \cite{saus_dataset}, \cite{sar_custom_drone_dataset}, \cite{merged-thesis_dataset}). \textcolor{black}{To construct the training and validation sets of our custom dataset, we used the original training and validation splits from four of these datasets \cite{forest-fire-and-smoke-rf4pd_dataset, saus_dataset, sar_custom_drone_dataset, merged-thesis_dataset} without applying any selection or filtering. From the dataset in \cite{Kularatne2024FireManPaper}, we selectively used only the Unimodal-RGB subset from the Multimodal collection, excluding the thermal modality. Furthermore, we included only a curated set of video frames that contained at least one of the following objects: fire, smoke, humans, or lakes. The resulting custom dataset comprises 40,384 training images and 11,953 validation images. The validation set was formed by concatenating the official validation splits from the five merged datasets rather than selecting a separate subset. This setup ensures a diverse, yet intentionally unbalanced distribution for evaluation purposes. The class imbalance provides a challenging scenario for evaluating fire, smoke, human presence, and water body detection, fostering robust model training under diverse conditions.} \textcolor{black}{We also evaluate E-IRFS on two well-known balanced datasets, CIFAR-10 \cite{krizhevsky2009cifar10} and Caltech-101 \cite{Fei2006Caltech101}, to assess its behavior when class distributions are uniform. This allows us to verify whether the exponential reweighting introduces performance degradation outside long-tailed scenarios.}

\begin{table}[ht!]
\vspace{-2mm}
\setlength{\tabcolsep}{0.3em}
\def\arraystretch{1.2}
\centering
\caption{Class Distribution in Training and Validation Sets in our custom dataset, Illustrating Class Imbalance.}
\vspace{-2mm}
\begin{tabular}{|c|c|c|c|c|c|c|}
\hline
 & \multicolumn{3}{|c|}{\textbf{Training Set}}  & \multicolumn{3}{|c|}{\textbf{Validation Set}}\\
\cline{2-4}
\cline{5-7}
\textbf{Class} & \textbf{Images} & \textbf{Instances} & \textbf{Percentage} & \textbf{Images} & \textbf{Instances} & \textbf{Percentage} \\
\hline
Fire   & 16,915 & 33,773 & 23.0\% & 1,436 & 2,336 & 8.5\% \\
Smoke  & 28,769 & 32,538 & 22.1\% & 6,735 & 7,090 & 25.8\% \\
Human  & 18,525 & 67,992 & 46.3\% & 4,804 & 16,612 & 60.5\% \\
Lake   & 12,646 & 12,646 & 8.6\% & 1,087 & 1,426 & 5.2\% \\
\hline
\textbf{Total} & \textbf{40,384} & \textbf{146,949} & \textbf{100\%} & \textbf{11,953} & \textbf{27,464}  & \textbf{100\%} \\ 
\hline
\end{tabular}
\end{table}

\subsection{Evaluation Metrics and Protocols}

The effectiveness of the YOLOv11 object detection models was evaluated using standard metrics provided by the Ultralytics validation system. These metrics assess the accuracy and reliability of the model in identifying objects across different conditions. To quantify detection performance, we used mean Average Precision at an Intersection over Union (IoU) threshold of 0.5 (mAP$_{50}$) and mean Average Precision across multiple IoU thresholds ranging from 0.5 to 0.95 in steps of 0.05 (mAP$_{50-95}$). The mAP$_{50}$ metric evaluates detection performance by computing the precision-recall curve at a fixed IoU threshold of 0.5, allowing for some localization tolerance. In contrast, mAP$_{50-95}$ provides a more strict assessment by averaging precision across multiple IoU thresholds, thereby offering a more comprehensive measure of detection accuracy and localization robustness. These metrics facilitate a standardized comparison of model performance across different object scales and levels of complexity within the dataset, ensuring a rigorous evaluation of the proposed approach. \textcolor{black}{For classification tasks on CIFAR-10 and Caltech-101, we report Top-1 and Top-5 accuracy, which measure whether the correct class is the model first or among its five highest-ranked predictions. These metrics are standard for evaluating classification without localization.}

%
%
\section{Experimental Analysis and Results} \label{sec: Experimental Analysis and Results}

\subsection{Experimental Setup}

All experiments used the Ultralytics YOLOv11 framework \cite{yolo11_ultralytics}. The YOLOv11-Nano model (2.6M parameters) and YOLOv11-Large (25.3M parameters) were trained on the custom dataset using an NVIDIA A30 GPU (24GB VRAM) to compare the impact of model size on performance. Training was performed using an image resolution of 640 × 640 pixels, batch size 16, and 100 epochs. The optimizer was auto-selected with momentum 0.937, weight decay 0.0005, and an initial learning rate of 0.01, gradually reduced to 0.001. Data augmentation was applied in all sampling-based experiments unless explicitly omitted, including Mosaic (0.9), Mixup (0.2), Copy-Paste (0.3), AutoAugment, and geometric transformations. Model performance was assessed using mAP$_{50}$ and mAP$_{50-95}$ metrics at different threshold values and values of the scaling parameter. The evaluation included comparisons between RFS, IRFS, and the proposed E-IRFS method, as well as baseline experiments with default hyperparameters and augmentation configurations.

\begin{figure*}[ht!]
    \centering
    \includegraphics[width=\linewidth, trim=0 200 0 100]{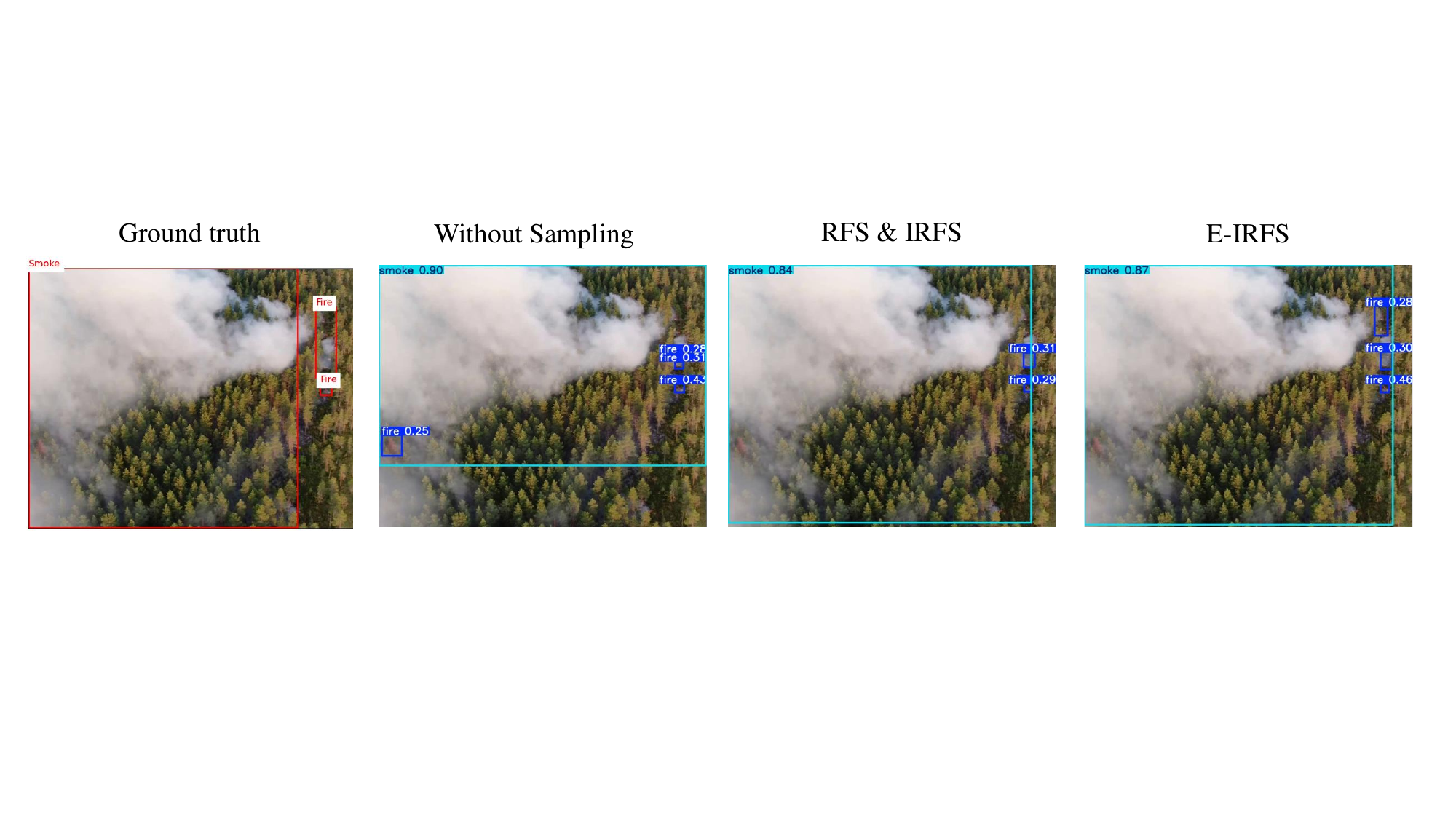} 
    \caption{Comparison of rare category detection performance on our custom testset for YOLOv11-Nano model.}
    \label{fig:myfigure}
\end{figure*}

\subsection{Performance Comparison of Re-Balancing Methods}

This experiment compares the RFS, IRFS, and E-IRFS with a baseline in the YOLOv11-Nano setup to evaluate their impact on rare-class detection and overall model performance with and without data augmentation. During training, we set \(\alpha =2.0 \) for E-IRFS, and threshold $t=0.0001$ for all methods. The results presented in Table \ref{tab:rebalancing_performance}, firstly, highlight the performance of re-balancing methods without the influence of data augmentation. Both RFS and IRFS improve mAP$_{50}$ by 8\% compared to the baseline but do not affect the mAP$_{50-95}$ metric. In contrast, E-IRFS achieves a 17\% increase in mAP$_{50}$ and a 13\% improvement in mAP$_{50-95}$, demonstrating its stronger ability to improve detection accuracy across different object sizes. The significant gain inmAP$_{50-95}$ suggests that E-IRFS not only enhances class balance but also contributes to better localization precision for rare objects.

\begin{table}[ht!]
\vspace{-2mm}
\setlength{\tabcolsep}{0.5em}
\def\arraystretch{1.2}
\centering
\caption{Comparison of mAP$_{50}$ and mAP$_{50-95}$ in the YOLOv11-Nano setup, trained with and without data augmentation.}
\vspace{-2mm}
\begin{tabular}{|c|c|c|c|c|c|c|}
\hline

& \multicolumn{2}{|c|}{\textbf{No Data Augmentation}}  & \multicolumn{2}{|c|}{\textbf{Data Augmentation}}\\
\cline{2-3}
\cline{4-5}

\textbf{Method} & \textbf{mAP$_{50}$} & \textbf{mAP$_{50-95}$} & \textbf{mAP$_{50}$} & \textbf{mAP$_{50-95}$}\\
\hline
Baseline  & 0.45 & 0.22 & 0.49 (+8\%) & 0.23 (+4\%) \\
RFS  & 0.49 (+8\%) & 0.22 (+0\%) & 0.50 (+11\%) & 0.24 (+9\%) \\
IRFS  & 0.49 (+8\%) & 0.22 (+0\%) & 0.50 (+11\%) & 0.24 (+9\%)  \\
E-IRFS  & 0.53 (+17\%) & 0.25 (+13\%) & 0.55 (+22\%) & 0.25 (+13\%) \\
\hline
\end{tabular}
\label{tab:rebalancing_performance}
\end{table}

To further assess the effect of re-balancing with data augmentation, Table \ref{tab:rebalancing_performance} presents results with augmentation applied across all methods, including the baseline. While RFS and IRFS improve mAP${50}$ by 11\%, their gains remain marginal compared to training without augmentation. In contrast, E-IRFS achieves a 22\% increase in mAP${50}$ and a 13\% improvement in mAP$_{50-95}$, further demonstrating its advantage in addressing class imbalance, as shown in Figure \ref{fig:myfigure}. These results suggest that while augmentation benefits all methods, E-IRFS benefits the most due to its exponential weighting, which amplifies the effect of balanced sample exposure. The performance differences highlight a key limitation of RFS and IRFS, where their linear weighting does not sufficiently enhance rare-class representation.

\subsection{Threshold and Scaling Factor Analysis in E-IRFS}

This experiment analyzes the impact of the threshold parameter and the scaling factor \(\alpha\) on the E-IRFS method, evaluating their effect on model performance. The formulation in Equation \ref{eq:sampprob} demonstrates that \(\alpha\) controls the aggressiveness of exponential scaling, while the threshold $t$ determines the starting point for rebalancing adjustments. By systematically varying these parameters, we assess their influence on model accuracy. The results in Table \ref{tab:eirfs_nano} and Table \ref{tab:eirfs_large} show that increasing \(\alpha\) improves the emphasis on rare categories, but only when combined with a sufficiently low threshold. 

\begin{table}[ht!]
\setlength{\tabcolsep}{0.3em}
\def\arraystretch{1.2}
\centering
\caption{mAP$_{50}$ of E-IRFS on YOLOv11-Nano with varying \(\alpha\) and threshold values}
\vspace{-2mm}
\begin{tabular}{|c|c|c|c|c|}
\hline
\textbf{\(\alpha\) \textbackslash Threshold} & \textbf{0.1} & \textbf{0.01} & \textbf{0.001} & \textbf{0.0001} \\
\hline
0.5  & 0.53 (+17\%) & 0.49 (+8\%) & 0.53 (+17\%) & 0.48 (+6\%) \\
1.0  & 0.52 (+15\%) & 0.49 (+8\%) & 0.52 (+15\%) & 0.49 (+8\%) \\
2.0  & 0.48 (+7\%) & 0.50 (+11\%) & 0.47 (+4\%) & \textbf{0.55 (+22\%)} \\
\hline
\end{tabular}
\label{tab:eirfs_nano}
\end{table}

For example, for the small model YOLOv11-Nano (Table \ref{tab:eirfs_nano}), setting \(\alpha =2.0 \) with $t=0.0001$ produces a 22\% improvement in mAP$_{50}$, outperforming lower \(\alpha\) values where the effect of exponential scaling is weaker. A similar trend is observed in YOLOv11-Large, where the same configuration leads to an increase 18\%. Lower values of \(\alpha\), such as 0.5 or 1.0, result in more conservative adjustments, which reduce the ability to compensate for extreme class imbalances. The improvement at \(\alpha =2.0 \) and a low threshold is due to the exponential function in Equation \ref{eq:sampprob}. When the geometric mean is small, indicating a rare class, exponentiation amplifies the repeat factor more significantly, ensuring stronger sampling weights and better representation in training. 

\begin{table}[ht!]
\setlength{\tabcolsep}{0.3em}
\def\arraystretch{1.2}
\centering
\caption{mAP$_{50}$ of E-IRFS on YOLOv11-Large with varying \(\alpha\) and threshold values}
\vspace{-2mm}
\begin{tabular}{|c|c|c|c|c|}
\hline
\textbf{\(\alpha\) \textbackslash Threshold} & \textbf{0.1} & \textbf{0.01} & \textbf{0.001} & \textbf{0.0001} \\
\hline
0.5  & 0.52 (+4\%) & 0.56 (+12\%) & 0.57 (+14\%) & 0.54 (+8\%) \\
1.0  & 0.55 (+10\%) & 0.58 (+16\%) & 0.53 (+6\%) & 0.57 (+14\%) \\
2.0  & 0.55 (+10\%) & 0.54 (+8\%) & 0.57 (+14\%) & \textbf{0.59 (+18\%)} \\
\hline
\end{tabular}
\label{tab:eirfs_large}
\end{table}

To evaluate the effect of threshold selection in E-IRFS against RFS and IRFS, we fixed \(\alpha = 2.0 \) and varied the threshold across different values, comparing the results with traditional sampling strategies. The results in Table \ref{tab:rfs_irfs} show that RFS and IRFS consistently achieve an 11\% increase in mAP$_{50}$ regardless of the threshold, indicating that their sampling adjustments remain unaffected by changes in this parameter.

In contrast, E-IRFS is highly sensitive to threshold selection, with performance improving as the threshold decreases. This sensitivity arises from its exponential weighting mechanism, which amplifies rare-class sampling more dynamically than the linear adjustments in RFS and IRFS. Lower thresholds enhance differentiation between class frequencies, improving rare-class representation without excessive oversampling. As shown in Table \ref{tab:eirfs_nano}, E-IRFS achieves its highest mAP$_{50}$ at $t=0.0001$, where the exponential function optimally balances sampling rates. At higher thresholds, sampling becomes more uniform, reducing rebalancing benefits and causing performance fluctuations. This highlights the importance of carefully selecting \(\alpha \) and $t$ for long-tailed object detection.

\begin{table}[h!]
\vspace{-2mm}
\setlength{\tabcolsep}{0.95em}
\def\arraystretch{1.2}
\centering
\caption{Comparison of mAP$_{50}$ for RFS, IRFS, and E-IRFS with \(\alpha =2.0 \) at different threshold values on YOLOv11-Nano.}
\vspace{-2mm}
\begin{tabular}{|c|c|c|c|c|c|}
\hline
\textbf{Threshold} & \textbf{RFS} & \textbf{IRFS} & \textbf{E-IRFS (\(\alpha =2.0 \))} \\
\hline
0.1  & 0.50 (+11\%) & 0.50 (+11\%) & 0.48 (+7\%)\\
0.01  & 0.50 (+11\%) & 0.50 (+11\%) & 0.50 (+11\%) \\
0.001  & 0.50 (+11\%) & 0.50 (+11\%) & 0.47 (+4\%) \\
0.0001 & 0.50 (+11\%) & 0.50 (+11\%) & 0.55 (+22\%)\\
\hline
\end{tabular}
\label{tab:rfs_irfs}
\end{table}


\subsection{Impact of Re-Balancing Methods on Class-Wise Performance}

The evaluation of re-balancing methods at the class level provides a more precise insights into their effectiveness in improving detection across underrepresented categories. The results in Tables \ref{tab:classwise_map50} and \ref{tab:classwise_map5095} indicate that E-IRFS achieves the most significant improvements, particularly for minority classes such as Lake and Fire.

\begin{table}[ht!]
\vspace{-1mm}
\setlength{\tabcolsep}{0.2em}
\def\arraystretch{1.3}
\centering
\caption{Class-Wise mAP-50 Performance in YOLOv11-Nano Setup}
\vspace{-2mm}
\begin{tabular}{|c|c|c|c|c|c|c|}
\hline
\textbf{Class} & \textbf{Baseline} & \textbf{Aug} & \textbf{Aug+RFS} & \textbf{Aug+IRFS} & \textbf{Aug+E-IRFS}\\
\hline
Fire & 0.32 & 0.43(+34\%) & 0.44(+37\%) & 0.44(+37\%) & 0.53(65\%) \\
Smoke  & 0.80 &0.81(+1\%) & 0.82(+2\%) & 0.82(+2\%) & 0.82(+2\%) \\
Human  & 0.67 &0.65(-3\%) & 0.67(+0\%) & 0.67(+0\%) & 0.78(+16\%) \\
Lake  & 0.02 &0.05(+150\%) & 0.06(+200\%) & 0.06(+200\%) & 0.09(+350\%) \\
\hline
\end{tabular}
\label{tab:classwise_map50}
\end{table}

For the Lake category, mAP${50}$ increased from 0.02 to 0.09, corresponding to a relative improvement of 350\%, while mAP${50-95}$ increased from 0.009 to 0.032, reflecting a gain of 255\%. Similarly, the Fire class showed an increase of 65\% in mAP${50}$ and 31\% in mAP${50-95}$. The Human class benefited from an increase 16\% in mAP${50}$ and an 18\% increase in mAP${50-95}$, indicating balanced improvements. In contrast, the Smoke class exhibited only marginal improvements, suggesting that it was already well-represented in the dataset.

\begin{table}[h!]
\vspace{-3mm}
\setlength{\tabcolsep}{0.2em}
\def\arraystretch{1.3}
\centering
\caption{Class-Wise mAP(50-95) Performance in YOLOv11-Nano Setup}
\vspace{-2mm}
\begin{tabular}{|c|c|c|c|c|c|c|}
\hline
\textbf{Class} & \textbf{Baseline} & \textbf{Aug} & \textbf{Aug+RFS} & \textbf{Aug+IRFS} & \textbf{Aug+E-IRFS}\\
\hline
Fire & 0.165 & 0.191(+15\%) & 0.213(+29\%) & 0.213(+29\%) & 0.217(31\%) \\
Smoke  & 0.387 &0.465(+20\%) & 0.456(+17\%) & 0.456(+17\%) & 0.404(+4\%) \\
Human  & 0.299 &0.282(-5\%) & 0.310(+3\%) & 0.310(+3\%) & 0.353(+18\%) \\
Lake  & 0.009 &0.012(+33\%) & 0.016(+77\%) & 0.016(+77\%) & 0.032(+255\%) \\
\hline
\end{tabular}
\label{tab:classwise_map5095}
\end{table}

These results confirm that E-IRFS effectively enhances model performance on underrepresented classes, making it more adaptable to long-tailed distributions. Unlike RFS and IRFS, which provide uniform improvements across categories, E-IRFS dynamically adjusts sampling weights, leading to a more pronounced effect on classes with fewer instances. This supports its suitability for applications where rare object detection is critical.

\subsection{\textcolor{black}{Generalization to Balanced Datasets}}

\textcolor{black}{To evaluate the applicability of E-IRFS beyond long-tailed scenarios, we tested it on two balanced datasets: CIFAR-10 \cite{krizhevsky2009cifar10} and Caltech-101 \cite{Fei2006Caltech101}. These experiments aim to determine whether the exponential weighting mechanism introduces performance degradation or bias when class distributions are approximately uniform. We trained the YOLOv11-Nano classification model on both datasets using baseline and E-IRFS configurations with \(\alpha =0.5 \), \(\alpha =1.0 \), and \(\alpha =2.0 \), fixing the threshold at $t=0.1$.}

\begin{table}[ht!]
\setlength{\tabcolsep}{0.65em}
\def\arraystretch{1.2}
\centering
\caption{Accuracy on Balanced Datasets (Top-1 / Top-5)}
\vspace{-2mm}
\begin{tabular}{|c|c|c|c|}
\hline
\textbf{Dataset} & \textbf{Method} & \textbf{Top-1 Acc} & \textbf{Top-5 Acc} \\
\hline
\multirow{4}{*}{CIFAR-10}
& Baseline & 0.802 & 0.989 \\
& E-IRFS (\(\alpha =0.5 \), $t=0.1$) & 0.802 & 0.989 \\
& E-IRFS (\(\alpha =1.0 \), $t=0.1$) & 0.802 & 0.989 \\
& E-IRFS (\(\alpha =2.0 \), $t=0.1$) & 0.802 & 0.989 \\
\hline
\multirow{4}{*}{Caltech-101}
& Baseline & 0.920 & 0.986 \\
& E-IRFS (\(\alpha =0.5 \), $t=0.1$) & 0.927 & 0.986 \\
& E-IRFS (\(\alpha =1.0 \), $t=0.1$) & 0.924 & 0.985 \\
& E-IRFS (\(\alpha =2.0 \), $t=0.1$) & 0.920 & 0.984 \\
\hline
\end{tabular}
\label{tab:balanced_datasets_performance}
\end{table}

\textcolor{black}{As shown in Table~\ref{tab:balanced_datasets_performance}, E-IRFS maintains performance parity with the baseline across all settings on CIFAR-10 and Caltech-101. Minor fluctuations in Caltech-101 are within acceptable bounds and do not suggest systematic degradation. These results confirm that E-IRFS does not introduce bias or degrade performance on balanced datasets, supporting its robustness and generalization beyond imbalanced scenarios.}

%
%
\section{Conclusion}

This paper introduces E-IRFS, a novel rebalancing strategy designed to improve rare-class detection in long-tailed object detection. The experimental results demonstrate that E-IRFS outperforms existing methods in detecting rare objects such as fires and water bodies in UAV-based monitoring. E-IRFS is especially beneficial for lightweight models with fewer parameters, such as YOLOv11-Nano, since they lack the capacity to learn long-tailed distributions effectively, making them more dependent on data sampling strategies. E-IRFS benefits from a lower threshold and a higher scaling factor, allowing more adaptive sampling adjustments. Beyond its empirical improvements, it provides a structured approach to addressing class imbalance, with differentiation between frequent and rare categories, while highlighting the importance of adaptive sampling strategies. \textcolor{black}{E-IRFS shows sensitivity to selection of \(\alpha \) and $t$, but our results indicate that effective values can generalize across datasets with minimal tuning. Future work may explore the extension of E-IRFS to other recognition tasks and the integration of it with additional augmentation techniques to improve its performance in highly imbalanced datasets. Furthermore, a critical direction involves investigating methodologies to determine the optimal \(\alpha \) and $t$ parameters based on statistical priors from new datasets. This would reduce the need for extensive manual tuning, thus enhancing the practicality and applicability of the method across diverse domains.}

\section*{Acknowledgment}
This research was supported by the Research Council of Finland 6G Flagship program (Grant Number 369116), FIREMAN project (Grant Number 348008), and Business Finland Neural Publish Subscribe for 6G Project (diary number 8754/31/2022).

\bibliographystyle{IEEEtran}
\bibliography{references}


\end{document}